\documentclass[sn-mathphys,Numbered]{sn-jnl}

\usepackage{graphicx}%
\usepackage{multirow}%
\usepackage{amsmath,amssymb,amsfonts}%
\usepackage{amsthm}%
\usepackage{mathrsfs}%
\usepackage[title]{appendix}%
\usepackage{xcolor}%
\usepackage{textcomp}%
\usepackage{manyfoot}%
\usepackage{booktabs}%
\usepackage{algorithm}%
\usepackage{algorithmicx}%
\usepackage{algpseudocode}%
\usepackage{listings}%
\usepackage{subfigure}%

\begin{document}

\title[Article Title]{SkelVIT: Consensus of Vision Transformers for a Lightweight Skeleton-Based Action Recognition System}


\author[1]{\fnm{Özge} \sur{Öztimur Karadağ}}\email{ozge.karadag@alanya.edu.tr}

\affil[1]{\orgdiv{Department of Computer Engineering}, \orgname{Alanya Alaaddin Keykubat University}, \orgaddress{\street{Alanya}, \state{Antalya}, \country{Turkey}}}

\abstract{Skeleton-based action recognition receives the attention of many researchers as it is robust to viewpoint and illumination changes, and its processing is much more efficient than the processing of video frames. With the emergence of deep learning models, it has become very popular to represent the skeleton data in pseudo-image form and apply Convolutional Neural Networks(CNN) for action recognition. Thereafter, studies concentrated on finding effective methods for forming pseudo-images. Recently, attention networks, more specifically transformers have provided promising results in various vision problems. 

In this study, the effectiveness of vision transformers (VIT) for skeleton-based action recognition is examined and its robustness on the pseudo-image representation scheme is investigated. To this end, a three-level architecture, SkelVit is proposed, which forms a set of pseudo images, applies a classifier on each of the representations, and combines their results to find the final action class. The performance of SkelVit is examined thoroughly via a set of experiments. First, the sensitivity of the system to representation is investigated by comparing it with two of the state-of-the-art pseudo-image representation methods. Then, the classifiers of SkelVit are realized in two experimental setups by CNNs and VITs, and their performances are compared. In the final experimental setup, the contribution of combining classifiers is examined by applying the model with a different number of classifiers. Experimental studies reveal that the proposed system with its lightweight representation scheme achieves better results than the state-of-the-art methods. It is also observed that the vision transformer is less sensitive to the initial pseudo-image representation compared to CNN. Nevertheless, even with the vision transformer, the recognition performance can be further improved by the consensus of classifiers.}


\maketitle

\section{Introduction}\label{sec1}

Human action recognition (HAR) has been widely studied for many years by computer vision researchers. HAR has diverse application areas such as visual surveillance, autonomous driving, robotics, and healthcare systems. Based on the data modality employed, the studies on this domain can be grouped into two categories. While a group of studies employs video data, that is a sequence of RGB data, another group of studies employs skeleton data. With the emergence of low-cost sensors, the modality of the skeleton has received the attention of many researchers as it brings in a much more compact representation, and it is not sensitive to context, background, illumination changes, etc. 

Skeleton-based action recognition methods can be categorized into three based on their representational scheme; vectorial representation, graphical representation, and pseudo-image representation.
 
In the Vectorial Representation scheme, the skeleton data is represented in vector form. Recurrent Neural Networks (RNNs) are the most commonly used approach that employs this representation due to their ability to process sequential data for classification. While the models in this method mainly depend on temporal data, the RNN architectures cannot model long-term dependencies. To overcome this problem, LSTM (Long Short Term Memory) \cite{LSTM}, which integrates self-connected memory cells to store information for longer durations, is proposed. Nevertheless, this group of approaches fails to model the spatial relations among skeleton parts. 

In the Graphical Representation approach, the skeleton is represented as a graph and actions are represented as a sequence of graphs. Graph Convolution Networks \cite{GCN} is the most widely used method for this purpose, and is combined with various methods in the literature \cite{Li_2019,Yan_2018,Shi_2019,GCN_review,Wu_2023}. Due to the representation scheme, this group of approaches heavily relies on modeling spatial data but fails to model temporal relations. While the relations among directly connected skeleton joints can be modeled, relations between indirectly connected skeleton joints can not be modeled effectively with this approach. 

In the Pseudo-Image Representation scheme action data consisting of a sequence of 3D skeleton information is represented in image form \cite{Du_2015, SkeleMotion,Liu_Liu,Liu_PR_2017,Ding_2017,Ke_2017,Chao_2018,Joze_2019,Li_2022}. In this way, Convolutional Neural Networks (CNN) can be directly employed on skeleton data for action recognition. In this approach, while forming the pseudo-image representation depending on the order of joints, and placement of the spatial and temporal data it is possible to obtain various representations for a sequence of skeleton data and it is observed that action recognition systems that are using the pseudo-image representation are sensitive to the image-formation scheme \cite{Jia_2020,SkeleMotion,skepxels}.


Recently the use of attention networks, specifically transformers, has provided promising results in various computer vision problems and they are being widely used in many problems 
\cite{AdaVit,Transformer_Survey,Bottleneck_Transformers,Hugo,SpectralFormer,StyleTr,Xin_2023}.  For the skeleton-based action recognition problem, the use of transformers \cite{transformer} has been investigated for the initial two representation approaches, that is with the Graph Convolutional Network (GCN) and  Recurrent Neural Network (RNN) models \cite{Plizzari,rvit}. However, it has not been employed for the pseudo-image representation scheme. The taxonomy of Xin et al. \cite{Xin_2023} also reveals that, in the literature, the pseudo-image representation is only employed with CNN in the skeleton-based action recognition domain. To the best of our knowledge, the use of transformers for the pseudo-image representation approach is not proposed. More specifically, the vision transformer (VIT) \cite{VIT} has not been employed on skeleton data for action recognition.

In this study, a representationally robust system for skeleton-based action recognition is proposed by combining a pseudo-image representation approach with the VIT model. First, the skeletal data is represented in a pseudo-image form, then the vision transformer is applied to recognize the underlying action class. The sensitivity of the VIT model on the representation scheme is analyzed experimentally by applying alternative pseudo-image formation styles and the performance of VIT is compared with the CNN model. Also, the effectiveness of an ensemble of classifiers is examined experimentally and the answer to the question 'Do we still need an ensemble of classifiers if we have a representationally robust classifier?' is investigated.

The main contributions of the study can be summarized in two folds;

\begin{itemize}
        \item A lightweight, yet representationally robust three level architecture is proposed for skeleton-based action recognition.  
        
        \item The sensitivity of classifiers on the pseudo-image formation schema is experimentally analyzed for the skeleton-based action recognition problem.
\end{itemize}

The study is organized as follows; skeleton-based action recognition studies that employ pseudo-image representation are reviewed in section \ref{sec:literature}, the proposed architecture, SkelVIT, is introduced in section \ref{sec:arch}, experiments are explained in section \ref{sec:experiments}, conclusions and future work is discussed in section \ref{sec:conclusion}.

\section{Literature on Skeleton-Based Action Recognition Methods}  
\label{sec:literature}
Skeleton-based action recognition methods with a pseudo-image representation scheme can take advantage of deep models originally developed for vision problems. However, there is a strong need to capture the  "optimal" representation, which leverages both the spatial and temporal relations embedded in the action data.  

In 2015 Du et al. \cite{Du_2015} proposed using CNN for action recognition encoding skeleton data in 3D form by representing temporal dynamics in columns and spatial structure in rows and in this way representing the video sequence of action in a single image. Similar to this study, the majority of the studies with this representation scheme, take the joints in an order based on their neighborhood relations. However, in many actions, joints without neighboring relations have common properties which are neglected in those studies. To overcome this disadvantage, Li et al. \cite{Chao_2018} proposed a hierarchical architecture, which learns features from joints and captures their co-occurrences using a CNN model. 


Wang et al. \cite{Wang_2016, Wang_2018} proposed representing the skeleton data by joint trajectory maps using HSB(hue, saturation and brightness) color space to encode spatial-temporal data in a sequence of skeleton data corresponding to an action. They obtain trajectory maps for each of the three Cartesian planes of the real-world coordinates of the camera and use coloring to represent motion on the trajectories. They employ a Convolutional Neural Network for each of the three trajectories and combine their scores to find the action label.

Li et al. \cite{Li_translation_scale_inv} proposed a translation-scale invariant image mapping by first grouping skeleton joints as groups of arm, leg, etc., and then relating xyz coordinates with the RGB color channels. Compared to the representation of Du et al. \cite{Du_2015}, they propose using more specific data while doing the normalization and in this way ensure scale invariance, i.e. minimum values are obtained for each sequence and each axis separately while doing the normalization. 

Liu et al \cite{Liu_PR_2017} treat skeleton joints as points in a 5D space, where the three dimensions are xyz coordinates, one dimension corresponds to the time stamp and the fifth dimension corresponds to the joint number. They select two of the five dimensions to convert the skeleton data of 5D to form a 2D image with three channels. By applying different dimension selections, they obtain a set of representations and employ CNN on those representations to obtain scores for action classes. As a final step, they employ weighted fusion to combine the results of CNNs.

Caetano et al. \cite{SkeleMotion} refers to the pseudo-image representation approach as a skeleton image, and proposes a new method for representation by including magnitude and orientation data for skeleton joints and proposes applying different temporal scales to be able to capture long-range dependencies. 

Li et a. \cite{Li_2022} proposed using three coordinates separately to obtain pseudo-images which are first processed by an attention module for spatial and temporal calibration, and then fused and processed by a CNN model.  

Pseudo-image representation is studied thoroughly and several alternatives have been proposed to overcome the disadvantages of CNN in terms of sensitivity to representation \cite{Wang_2016,SkeleMotion}. However, alternative methods which can be applied to the pseudo-image representation of skeleton data have not been proposed. However, CNN has the obvious disadvantage of dependence on local representations due to its bottom-up processing model which starts with convolution. On the other hand, vision transformers are based on modeling the relations among distant image parts. For this reason, in this study, a skeleton-based action recognition system employing vision transformers on the pseudo-image representation of skeleton data is proposed. The effectiveness of attention networks on skeleton-based action recognition using the pseudo-image representation scheme is investigated and the robustness of attention networks on the pseudo-image data representation is examined.

\section{SkelVit Architecture}
\label{sec:arch}

\begin{figure}
\centering
\includegraphics[scale=0.4]{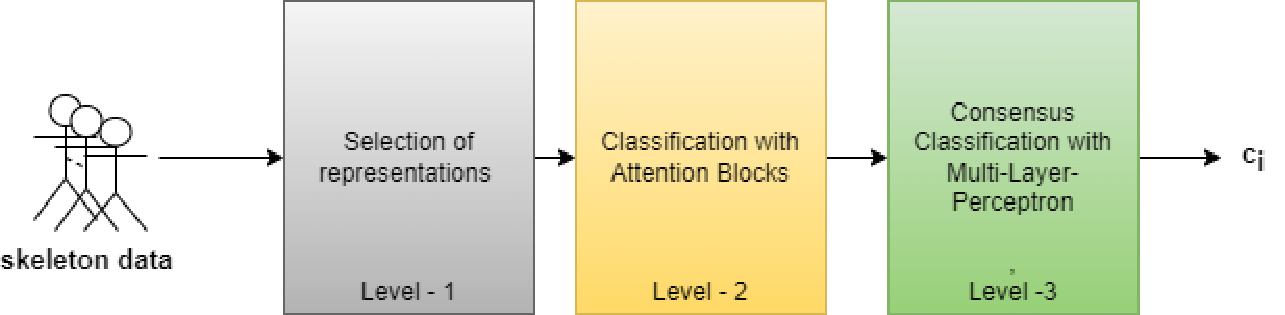}
\caption{Abstract view for the system Architecture of SkelVit.}
\label{fig:sys_arch_abs}
\end{figure}

In this section, we present a formal introduction to the proposed SkelVit architecture, a three-level framework designed for the recognition of actions, based on skeletal data. The first level, involves the representation of skeletal data in the form of  pseudo images, offering a versatile platform for multiple representations of these pseudo images. Subsequently, an algorithm is proposed to systematically select a set of potential representations based on this versatile framework. In the second level, an ensemble of classifiers is introduced, each trained to discern and select the optimal representation. In the third level, the decisions of the classifiers from the second level are aggregated to derive the final decision, thus culminating in a robust and effective mechanism for skeleton-based action recognition. The abstract view of the proposed system is shown in Figure~\ref{fig:sys_arch_abs}, and its details are explained in the subsequent sections.

\subsection{Level 1 - Generation and Selection of Pseudo Images} 
In the first part of level 1, we suggest a technique for the generation of the pseudo images from the skeletal data. Then, among all possible arrangements of pseudo-image representation, we select the "best" set of arrangements by using a robust and simple criterion. 

\subsubsection{Generation of Pseudo Images}

The skeletal data for an action consists of frames, each of which is indexed by  $t= 1,...,T$, where $T$ represents the total number of frames. Each frame  consists of a set of ordered joints $J_o^t=\{j_k^t\}_{k=1}^M$, where $o$ represents a pre-defined order and the $k$-th joint of frame $t$ is represented by the three-dimensional spatial coordinate vector, as follows:
\begin{equation}
     j_k^t=[x_k^t\quad y_k^t \quad z_k^t].
\end{equation}
In the above equation  $k \in [1,...,M]$ represents the joint index and $M$ represents the total number of joints in each frame $t \in [1,...,T]$. Hence, the skeleton data  $J_o^t=\{j_k^t\}_{k=1}^M$ with a pr-edefined order $o$ ,  at frame $t$ is represented by a sequence of three-dimensional spatial coordinates, corresponding to the joint information.

\begin{figure}
    \centering
    \subfigure[]{\includegraphics[width=0.24\textwidth]{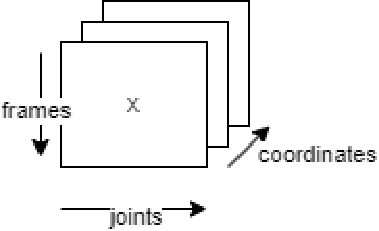}} 
    \subfigure[]{\includegraphics[width=0.18\textwidth]{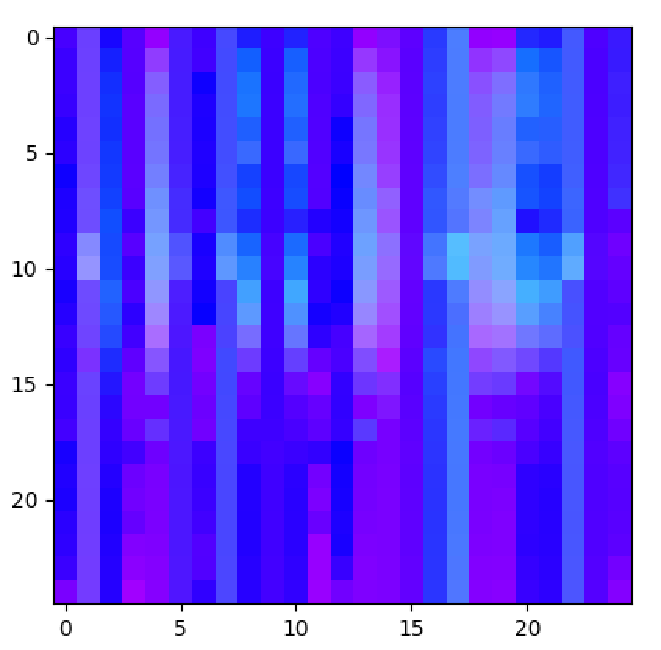}} 
    \subfigure[]{\includegraphics[width=0.18\textwidth]{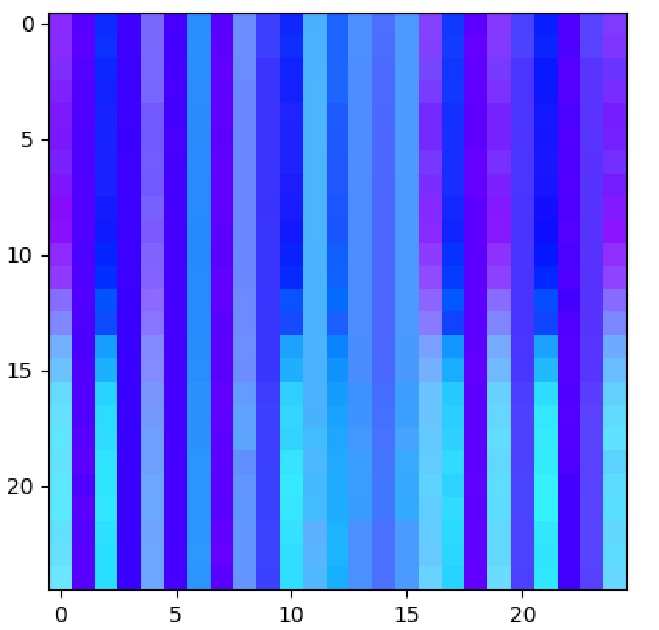}}
    \caption{Pseudo-image formation. a) Each frame $t$, consits of a set of joints, $J^t=\{j_k^t\}_{k=1}^M,$ where each joint consists of three dinensional coordinates,$j_k^t =[x_k^t y_k^t z_k^t]$ . (b) and (c) sample pseudo images, each represented by 8-bit/pixel RGB matrices, $F_x, F_y$ and $ F_z$. }
    \label{fig:image_representation}
\end{figure}

This joint information is used to construct a spatio-temporal representation in a pseudo-image form. 
This process is depicted in Figure \ref{fig:image_representation} (a). In this figure, the $x,y,z$ coordinate values are treated as the channels of a three-dimensional color space, such as RGB. The three channels of the pseudo-image are formed by the  $x, y$ and $z$- coordinate values of body joints, which are arranged based on a pre-defined order, $o$: 
$$f_t^x= [x_{o}^t \quad x_{o}^t\quad...\quad x_{o}^t]^\mathbb{T},$$
$$f_t^y= [y_{o}^t  \quad y_{o}^t \quad...\quad y_{o}^t]^\mathbb T,$$
$$f_t^z= [z_{o}^t  \quad z_{o}^t \quad ... \quad {o}^t]^{\mathbb T},$$
where $\mathbb T$ shows the vector transpose. As the number of joints is $M$, it is possible to define $M!$ different permutation  of joint orders. Each of the possible orders is referred to as a joint arrangement, $J_o$, where the index $o \in [0,...,M!]$ shows a specific joint arangement.

Then, for a predefined joint order $o$, the feature vectors of  all frames  are combined under matrices of  size $T \times M$, to obtain the $x, y$ and $z$ channels of the pseudo-image, as follows:
\begin{equation}
F_o^x = [f_1^x.......f_T^x],\quad
F_o^y = [f_1^y.......f_T^y], \quad
F_o^z = [f_1^z.......f_T^z]
\end{equation}

 Once all three channels are obtained, the feature matrices, $F_o^x, F_o^y$ and $F_o^z$ are scaled to range $[0,255]$ to obtain the 8 bits/pixel RGB pseudo-images. Sample images formed in this way are provided in Figure \ref{fig:image_representation} (b) and (c). 


The block diagram representation of Level 1 is given in Figure \ref{fig:sys_arch_lev1}. In this figure, first a set of possible joint arrangements, 
$$\mathbb{JA}=\{J_o\}_{o=1}^{M!}$$ are obtained for all possible joint orderings, $o \in [1,2,...,M!]$. Then, the $L$ number of these arrangements is randomly selected to obtain a subset, $\mathbb{JA}_n\subset\mathbb{JA}$, which is denoted as, $$\mathbb{JA}_{n}=\{J_{l}\}_{l=1}^L.$$
This subset is later used to form pseudo-images as described above. The random selection process is repeated $N$ times. Hence, $N$ many RGB matrices $F_o^x ,\space F_o^y$ and $F_o^z$ of size $L\times M$ are generated. 


\subsubsection{Selection of Pseudo Images} 
\begin{figure}
\centering
\includegraphics[scale=0.4]{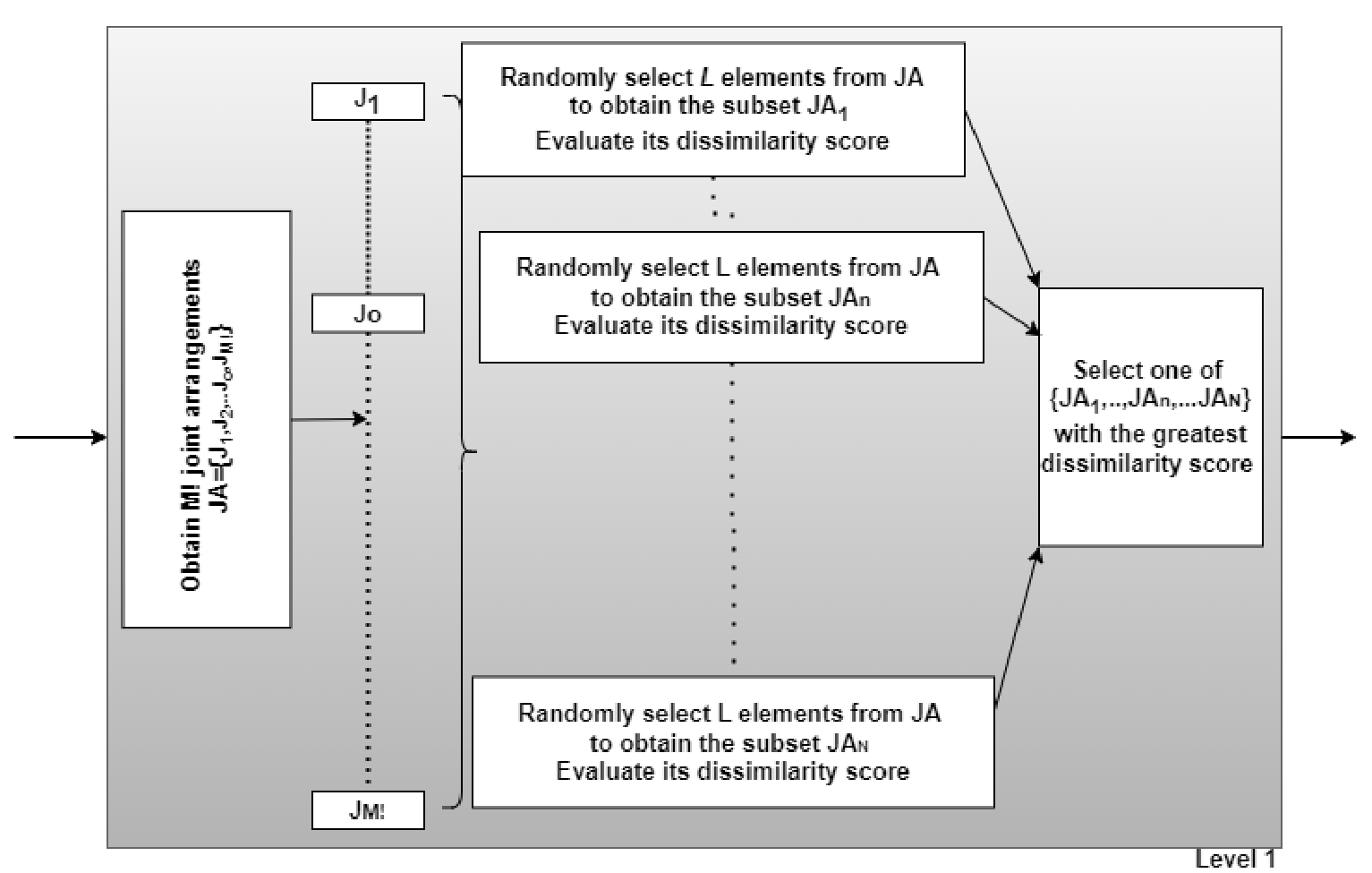}
\caption{System architecture for Level 1 of SkelVit.}
\label{fig:sys_arch_lev1}
\end{figure}

One of the subsets $\{\mathbb{JA}_n\}_{n=1}^N$ in Figure ~\ref{fig:sys_arch_lev1}, will be selected to be used in the next level to create pseudo-images. To take advantage of various orderings, in other words, to be able to represent various joint relations, it is important to combine dissimilar joint arrangements. Among the subsets, $\mathbb{JA}_n$, the subset with the highly scattered joints is considered to be a representationally more powerful matrix compared to matrices with a similar arrangement of joints. 

To select the representationally "most powerful" matrix a dissimilarity score is defined for each subset $\mathbb{JA}_n$, as follows:

\begin{equation}
    \Delta(\mathbb{JA}_n)=\sum_{l=1}^{L}\sum_{m=1}^{M} {\delta(j_m,J_l)},
\label{eq:dist_1}
\end{equation}
where

\begin{equation}
    \delta(j_m,J_l)=\sum_{q\neq l, q=1}^{L-1} |x-x_q|.
\label{eq:rad_dist}
\end{equation}

In  Equation ~\ref{eq:dist_1} and ~\ref{eq:rad_dist}, the subset $\mathbb{JA}_n\subset\mathbb{JA}$  is represented in matrix form, which consists of $L$ rows each corresponding to a different joint ordering from the set $\mathbb{JA}$. $J_l$ denotes the $l$-th element of set $\mathbb{JA}_n$, and the position of joint $j_m$ in $J_l$ is represented by $x$ and for each joint, the elements of $\mathbb{JA}_n$ are considered in pairs to calculate the dissimilarity score, $\Delta(\mathbb{JA}_n)$. 

Among $N$ subsets, the subset $\mathbb{JA}_n$ with the highest dissimilarity score is selected as the combination of different representations corresponding to $L$ different arrangements of joints.


 Contrary to theavailable skeleton-based action recognition methods, the proposed SkelVit approach forms several plain representations. As an example, the method suggested by Cao et al.  \cite{Cao_2019}, has a relatively more complex representation obtained by combining several possible pseudo-images for various joint traversals under a single schema. Similarly, in \cite{skepxels}  Liu et al. employ deep architectures in several streams to model spatial-temporal relations. Instead of combining these representations on a single schema, Li et.al. \cite{Li_2022}, employ consensus classification by incorporating the results of classifiers applied to each pseudo-image representation. In this study, owing to this plain representation scheme, a lightweight architecture is constructed using vision transformers, as explained in the following sections.

\subsection{Level 2 - Classification with Attention Blocks}
Each arrangement $J_l$ in the selected set $\mathbb{JA}_n$ has a corresponding pseudo-image representation, as shown in Figure \ref{fig:image_representation}. 

\begin{figure}
\centering
\includegraphics[scale=0.4]{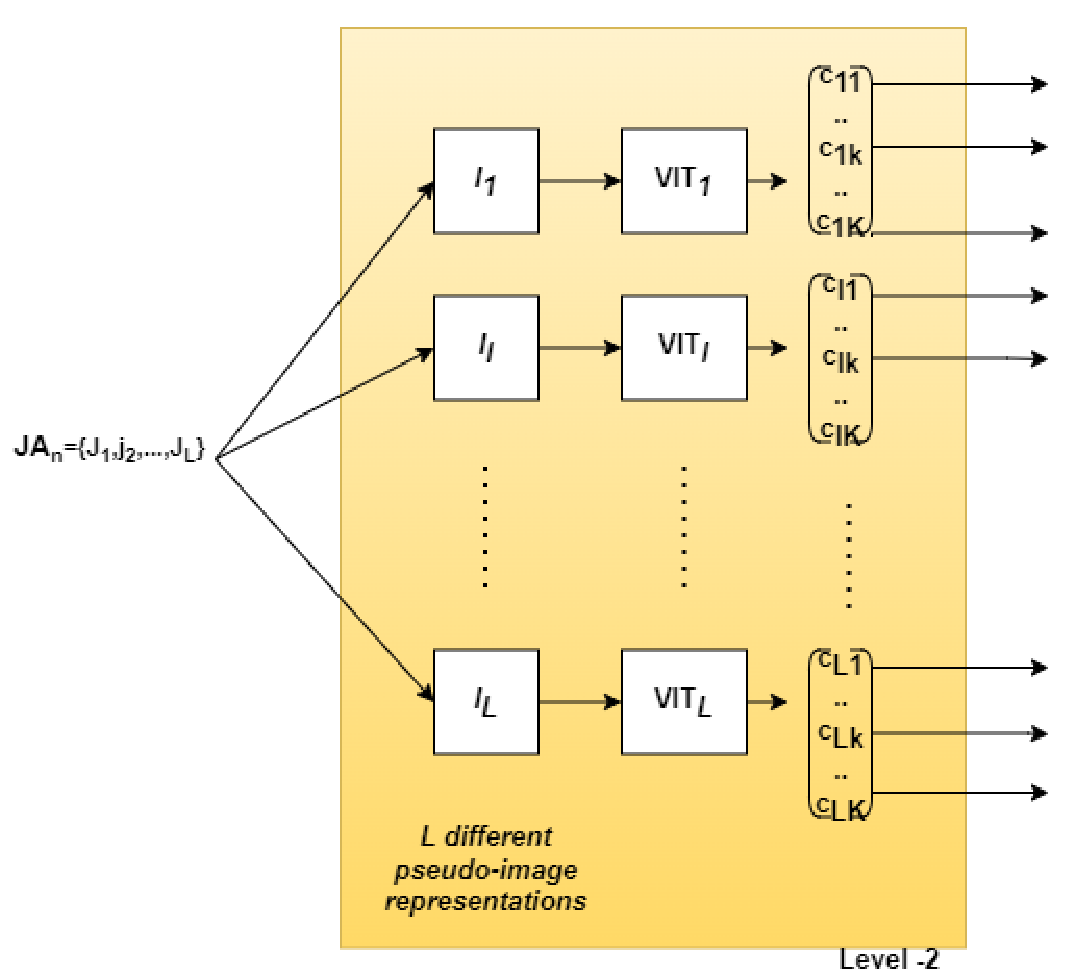}
\caption{System architecture for Level 2 of SkelVit.}
\label{fig:sys_arch_lev2}
\end{figure}

\begin{figure}
    \centering
    \includegraphics[scale=0.4]{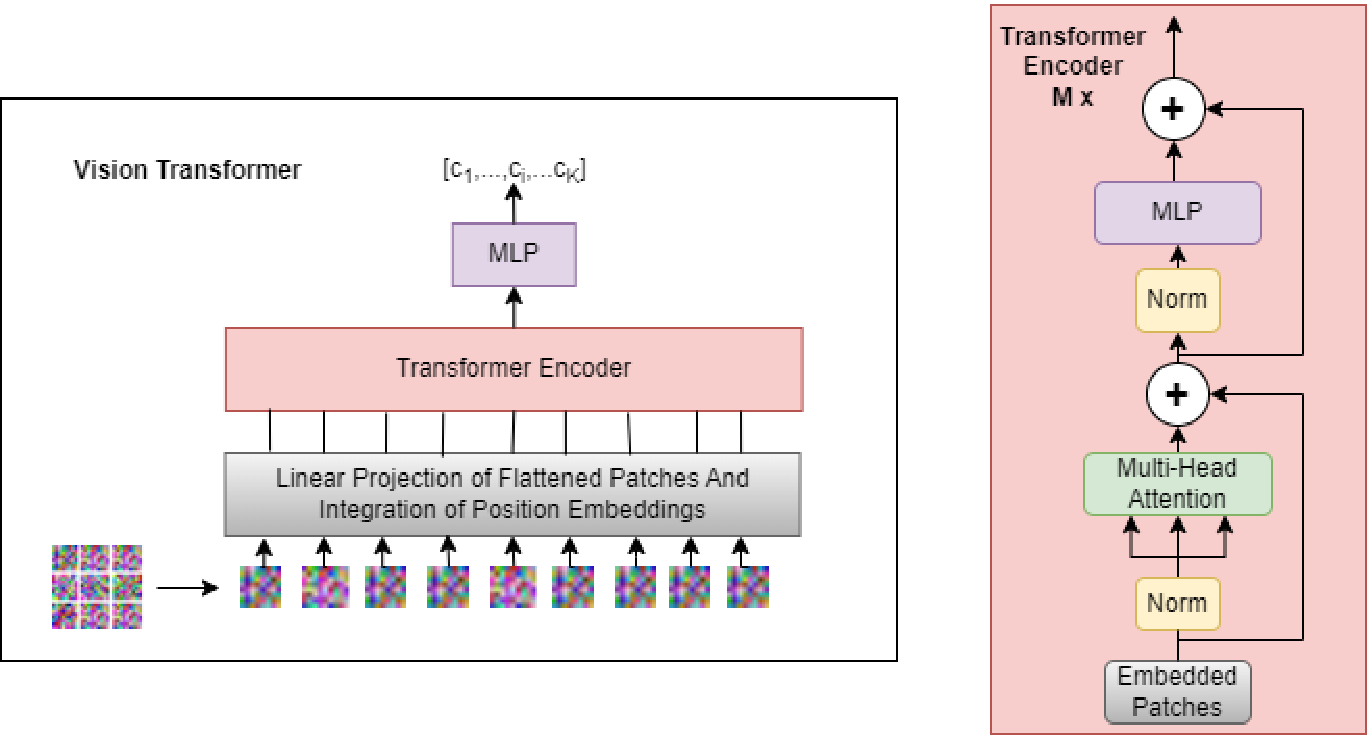}
    \caption{Vision Transformer, figure adapted from \cite{VIT}.}
    \label{fig:VIT}
\end{figure}

The block diagram of level 2 is given in Figure \ref{fig:sys_arch_lev2}. At level 2, each spatio-temporal representation, $I_l$ corresponding to $J_l$, which is obtained by combining the three vectors $F_l^x, F_l^y$, $F_l^z$, is processed by a Vision Transformer \cite{VIT} for action recognition. The architecture of VIT is shown in Figure \ref{fig:VIT}. Each VIT receives an image, divides the image into patches, adds position embeddings, and feeds the resulting vector into a transformer encoder. The transformer encoder processes the embedded patches in the Multi-Head Attention(MHA) module and evaluates query-key-value matrices as  Vaswani et al. \cite{transformer} suggested. Then the output of the MHA is processed by Multi-Layer Perceptron(MLP) as shown in Figure \ref{fig:VIT}. The transformer encoder consists of $M$ such parallel processing modules which are then combined and sent to a final MLP module which outputs class probabilities for all classes. 

\subsection{Level 3 - Consensus Classification by Multi-Layer Perceptron}
Level 3 receives class posterior probabilities from the VITs of level 2 and concatenates their results to be sent to a three-layered MLP network whose output gives the final class decision for SkelVit. The first layer of MLP consists of $L \times c$ number of neurons, the second layer is a hidden layer and the output layer consists of $c$ neurons.

\section{Experiments}
In this section, firstly, briefly the dataset, used in the experimental studies is explained. Secondly, the sensitivity of SkelVit to the representation scheme is experimentally analyzed by comparing it with state-of-the-art pseudo-image formation methods for skeleton-based action recognition. Thirdly, Levels 2 and 3 of the architecture are analyzed by comparing the classification accuracies of CNNs, VİTs and consensus of them separately. Finally, the effectiveness of consensus classification with varying numbers of classifiers is investigated. 
\label{sec:experiments}
\subsection{Dataset}
\begin{figure}
\centering
    \includegraphics[scale=0.5]{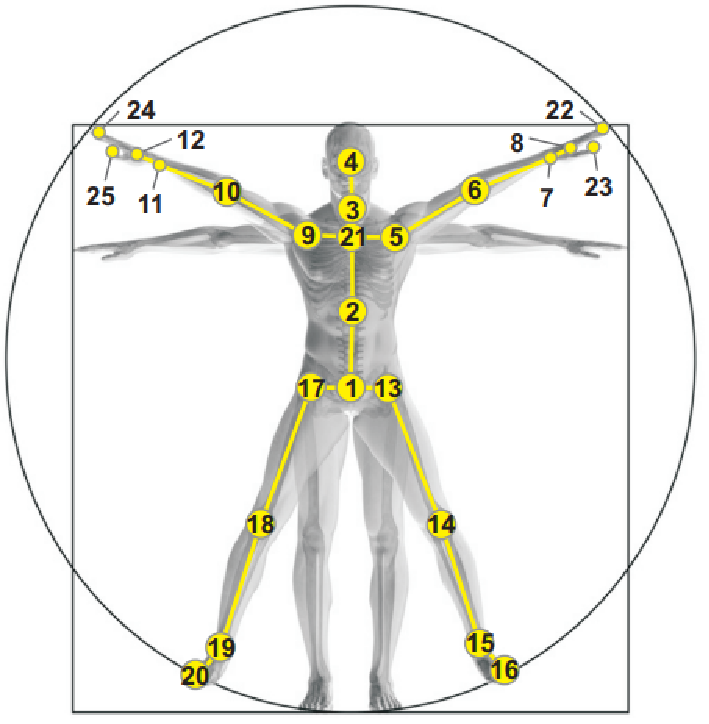}
    \caption{Configuration of 25 body joints in the NTU RGB+D dataset. Image is adapted from \cite{NTU}.}
    \label{fig:joints}
\end{figure}

In the experimental studies, a subset of NTU RGB+D Dataset is employed \cite{NTU}. The dataset contains $60$ classes including daily, mutual, and health-related actions. For each action, the dataset contains RGB video data, depth map sequences, 3D skeletal data, and infrared (IR) videos.  Each dataset is captured by three Kinect V2 cameras concurrently. In this study, among the data modalities, only the 3D skeletal data which contains the 3D coordinates of 25 body joints at each frame is employed. The joints are shown in Figure \ref{fig:joints}. 

Among  the daily actions, represented by the skeletal data, we select the classes, which are recognized with an accuracy higher than  $0.70$, as in \cite{Wang}. In the selected subset there are $c=14$ classes corresponding to; \textit{pick up, sit down, stand up, put on jacket, hand waving, take off jacket, put on a shoe, put on glasses, take off glasses, put on hat/cap, take off hat/cap, cheer up, hopping, jump up} action classes and there are a total of $13.191$ videos. The dataset is divided into train and test in two different ways in the literature. It is either divided as cross-subject (cs), where the train and test samples are obtained from different subjects, or it is divided as cross-view (cv), where the train and test samples are gathered from different cameras.

\subsection{Pseudo-image Representation}

In SkelVit each action is represented by a $T \times M$ pseudo-image, where both the number of frames, $T$, and the number of joints, $M$ are set to $25$. The pseudo-images are generated by the method, described in Figure \ref{fig:image_representation}. In the  experiments, a comprehensive set of twenty-five frames is derived for each action sample through the selection of one frame from every three consecutive frames. In this way, a compact representation, which is vital for the proposed lightweight architecture, is obtained without compromises. 

The effectiveness of the image representation of SkelVit is compared with state-of-the-art pseudo-image representation methods for action recognition. For this purpose, Level-1 of the architecture is realized by employing two state-of-the-art-methods; Skepxels \cite{skepxels} and Enhanced Skeleton Visualization \cite{Liu_PR_2017}. 
Skepxels, suggested by Liu et al. \cite{skepxels} map skeleton data into an image. Given a skeleton sequence, one may end up with different skepxels by employing different joint arrangements. They propose to concatenate several skepxels to obtain a larger image and use the image for action representation. For a given action, several such representations are obtained and a subset of them are selected by estimating the scatterings.

Enhanced Skeleton Visualization \cite{Liu_PR_2017} method proposes to arrange five-dimensional video data $(f,j,x,y,z)$, where the first two dimensions correspond to the spatial image coordinates and other three dimensions are used as $(r,g,b)$ components of the image. With five coordinates, ten different combinations can be obtained. At Level 1 of the architecture, all ten combinations are obtained, at Level 2 VITs are employed on each representation and their decisions are ensembled at Level 3. In this setup, with Enhanced Skeleton Visualization, no selection schema among different representations is employed.

The evaluation of the three methodologies for action recognition is conducted utilizing the accuracy metric, which evaluates the percentage of correctly classified samples. The obtained accuracies for cross-subject and cross-view evaluation are presented in Table \ref{tab:results_2}. It is discerned that SkelVit surpasses other contemporary approaches, when applied to a subset of the NTU-RGBD dataset. Although SkelVit draws inspiration from the Skepxels method in terms of architecture, it demonstrates a more promising outcome than Skepxels. Additionally, SkelVit proves to be more advantageous than Skepxels, owing to its compact representation. This representational scheme empowers SkelVit to employ a lightweight deep architecture at levels 2 and 3.


\begin{table}
    \centering
    \begin{tabular}{c|c|c|}
        Method & cs-accuracy & cv-accuracy \\
        \hline
         Enhanced Skeleton Visualization \cite{Liu_PR_2017} & 62.31 & 66.60  \\
         Skepxels \cite{skepxels} & 66.13 & 75.03 \\
         SkelVit & 73.44     & 80.85 \\
    \end{tabular}
   $ \\$
    \caption{Comparison of SkelVit with state-of-the-art methods on pseudo-image formation for action recognition via accuracy measure for cross-subject and cross-view evaluation.}
    \label{tab:results_2}
\end{table}

\subsection{Classification Accuracies}
\label{sec:classification}
The robustness of VIT architecture is experimentally evaluated by employing two different architectures on level 2 without changing the structure of level 1 and level 3. On the first setup, CNNs are employed for classification; on the second setup, VITs are employed instead of CNNs. The CNN architecture consists of two convolution-max pooling blocks, which are followed by a flatten, two dense, a dropout, and a final dense layers. The VIT architecture consists of eight transformer blocks each with four heads for attention. The patch size for the VIT is set as six. In both of the setups, the subset selection process is repeated $N=1000$ times, and the representation with the highest dissimilarity score is selected using Equation \ref{eq:dist_1}. The number of input neurons on level 3 is set as $25\times14$, as $L=25$ and $c=14$, the number of neurons at the hidden layer is set as $512$ and the number of neurons at the output layer is set as $c=14$. The number of epochs is set as $100$ and the learning rate is set as $0.001$ and the Adam optimizer is employed.  

The dataset is partitioned into two, setting the number of train samples as $8773$ and the number of test samples as $4418$. 

In the assessment of performance, four criteria are utilized: alongside classification accuracy,  precision is employed to quantify the number of correct positive predictions by $TP/(TP+FP)$; recall is utilized to quantify the number of correct positive predictions made out of all positive predictions that could have been made by $TP/(TP+FN)$; and the f-score, derived from precision and recall through the formula $F-Score=(2 \times precision \times recall)/(precision+recall)$ is used for evaluation.


Experimental results are summarized in Table \ref{tab:results}. In this table, results for both cross-subject and cross-view setups are given.  The results of the first experimental setup, where CNNs are employed in level 2, are reported in the second column as the average of CNNs and in the third column as the consensus of CNNs, and the results of the second experimental setup, where VITs are employed, are in the fourth column as the average of VITs and in the fifth column as the consensus of VITs. When the average performance values for CNNs and VITs are compared, it is observed that VIT outperforms CNN in skeleton-based action recognition using pseudo-image representation. What is more, Considering both models' robustness to representation, VIT is less sensitive than CNN to the initial representation. This is concluded from the fact that the performance improvement is higher when the consensus of CNNs is employed instead of averaging CNNs, than the performance improvement when the consensus of VITs is employed instead of averaging VITs. The accuracy of CNN increased from $64.50$ to $70.96$ for cs and it increased from $73.60$ to $79.96$ for cv, similarly f-score of CNN increased from $62.98$ to $69.19$ for cs and it increased from $71.65$ to $78.26$ for cv. On the other hand, when the consensus of VITs is used instead of the average of VITs, the accuracy increases from $70.29$ to $73.43$ for cs setup and it increases from $77.62$ to $80.85$ for cv setup. So it is noteworthy to point out that the contribution of combining classifiers is more explicit for CNN classifiers than VIT classifiers. As the main difference between the set of classifiers in level 2 is the representation, it is concluded that the VIT is less sensitive to the initial representation than the CNN. Nevertheless, still, the different representations have their advantages in representing certain actions, and their combination improves the action recognition performance.

\begin{table}[]
    \begin{tabular}{@{}lllllllll@{}}
    \cmidrule(l){2-9}
                               & \multicolumn{2}{l}{Average of CNN} & \multicolumn{2}{l}{Consensus of CNN} & \multicolumn{2}{l}{Average of VIT} & \multicolumn{2}{l}{Consensus of VIT} \\ \cmidrule(l){2-9} 
                                    & cs     & cv                        & cs      & cv                         & cs    & cv                         & cs                & cv               \\ \midrule
    \multicolumn{1}{l|}{accuracy}  &    64.50    & \multicolumn{1}{l|}{73.60}     &   70.96      & \multicolumn{1}{l|}{79.96}      & 70.29 & \multicolumn{1}{l|}{77.62} & \textbf{73.43}   & \textbf{80.85}            \\
    \multicolumn{1}{l|}{precision} &   61.48     & \multicolumn{1}{l|}{69.80}     &   67.48      & \multicolumn{1}{l|}{76.63}      & 67.20 & \multicolumn{1}{l|}{74.16} & \textbf{70.82}   & \textbf{77.76}            \\
    \multicolumn{1}{l|}{recall}    &   64.57     & \multicolumn{1}{l|}{73.62}     &   71.00      & \multicolumn{1}{l|}{79.97}      & 70.34 & \multicolumn{1}{l|}{77.63} & \textbf{73.48}   & \textbf{80.86}            \\
    \multicolumn{1}{l|}{f-score}   &  62.98      & \multicolumn{1}{l|}{71.65}     &  69.19       & \multicolumn{1}{l|}{78.26}      & 68.73 & \multicolumn{1}{l|}{75.85} & \textbf{72.12}    &  \textbf{79.27}                  \\ \bottomrule
    \end{tabular}
    \caption{Comparison of the CNN and VIT models.}
    \label{tab:results}
\end{table}


 
\subsection{Ablation Studies}
In the previous two subsections, SkelVit is investigated thoroughly to reveal the importance of both the choice of the representation scheme and the classifier. To this end, the representation method is compared with state-of-the-art representation models of pseudo-image representation for skeleton-based action recognition. Moreover, level 2 is realized both by VITs and CNNs to examine the contribution of both classifiers. In this section, SkelVit is further investigated for its robustness to parameters. Its sensitivity to the number of classifiers, $L$ in level 2, is examined. For this purpose, in three separate experimental setups, setting the other parameters as in section \ref{sec:classification}, SkelVit is realized by setting $L$ as ${10,25,40}$ respectively. The corresponding accuracies are given in table \ref{Table:L}.

\begin{table}[]
\begin{tabular}{llll}
\hline
                          & \begin{tabular}[c]{@{}l@{}}num. of classifiers\\ (L)\end{tabular} & cs    & cv    \\ \hline
\multirow{3}{*}{Accuracy} & 10                                                                & 72.08 & 80.82 \\  
                          & 25                                                                & 73.43 & 80.85 \\  
                          & 40                                                                & 73.57 & 80.94 \\ \hline
\end{tabular}
\caption{SkelVit accuracies for cs and cv setups, for the number of classifiers, L, set as 10, 25 and 40.}
\label{Table:L}
\end{table}

It is observed that the contribution of combining classifiers is improved with the increase in the number of classifiers to an extent. However, as the number of classifiers increases, the complexity of the system increases as well. For this reason, $L=25$ can be considered as the optimum parameter setting for this problem.

\section{Conclusion}
\label{sec:conclusion}
Recently, deep models have almost become the key standard for various machine learning problems, ranging from robotics to health informatics. Convolutional neural networks have been very popular in vision problems and their power is adapted to other domains by establishing pseudo-image representations. Skeleton-based action recognition is one of those areas, which employs CNN architectures on the pseudo-image representation of skeleton data. This approach proposes a computationally efficient solution for the skeleton-based action recognition problem and takes advantage of the computational power of the deep models. Previous studies reveal that the recognition performance is sensitive to the initial pseudo-image formation procedure, when it is feeded to a CNN model. 

In this study, first, the suggested pseudo-image formation method  is compared with two state-of-the-art pseudo-image formation methods. Skepxels, which is based on revealing representations for different joint arrangements and Enhanced Skeleton Visualization, which is based on pseudo-image formation for different combinations of joints, frames and x-y-z coordinates. It is experimentally observed that Skepxels and SkelVit provide better results compared to the Enhanced Skeleton Visualizations. Also, it is observed that, despite its compactness, SkelVit is able to outperform the two methods mentioned above. 

Second, the sensitivity of VIT on the initial representation is investigated and compared to that of the CNN models. The contribution of consensus of classifiers on the robustness of representation is also examined for the skeleton-based action recognition problem.  To this end, a three-level architecture which uses a set of representations and employs consensus of classifiers, is proposed. On the first level, a set of pseudo-images is obtained and a group of them which are less alike each other is selected. On the second level, a classifier is applied to each of the pseudo-images to obtain action scores. On the third level, the posterior probabilities from the second level are ensembled by an MLP network to obtain final action labels. The proposed architecture is realized with two different setups, on the first setup CNN is employed, and on the second setup VIT is employed as a classifier on the second level of the architecture. Through experimental studies, the performance of CNN, VIT and consensus of CNNs and consensus of VITs are examined and the sensitivity of CNN and VIT on the initial representation is monitored. It is observed that VIT is less sensitive to the pseudo-image formation method compared to CNN. VIT, using key-query-value processing, tries to reveal the dependencies in the image and meanwhile obtains the class label of the image. This processing scheme makes it less dependent on local relations, and hence, less sensitive to the pseudo-image formation.

In this study, the robustness of vision transformers on the initial representation is validated.  The proposed lightweight architecture of consensus classifiers  can be employed in other domains and the effectiveness of ensembling the attention architectures can be further investigated either on pseudo-images  or in vectorial data.


\section{Declarations}

\textbf{Ethical Approval} 

This declaration is not applicable.

\begin{flushleft}\textbf{Funding} \end{flushleft}

No funding was received for conducting this study. The authors have no relevant financial or non-financial interests to disclose.

\begin{flushleft}\textbf{Availability of data and materials}\end{flushleft}

The NTU-RGBD dataset \cite{NTU} that is used in the experiments of this study is publicly available at https://rose1.ntu.edu.sg/dataset/actionRecognition/.

\end{document}